%% file: acl_latex.tex
\pdfoutput=1
\documentclass[11pt]{article}

\usepackage[]{acl}

\usepackage{times}
\usepackage{latexsym}

\usepackage[T1]{fontenc}
\usepackage[utf8]{inputenc}

\usepackage{microtype}

\usepackage{inconsolata}
\usepackage{arydshln}

\usepackage{url}            % simple URL typesetting
\usepackage{booktabs}       % professional-quality tables
\usepackage{amsfonts}       % blackboard math symbols
\usepackage{nicefrac}       % compact symbols for 1/2, etc.
\usepackage{microtype}      % microtypography
\usepackage{xcolor}         % colors
\usepackage[export]{adjustbox}
\usepackage{dashrule}
\usepackage{multirow}
\usepackage{graphicx}
\usepackage{amsmath}
\usepackage{amssymb}
\usepackage{amsfonts}

\def\ModelName{\texttt{COMEDY}}
\def\Dataset{\textbf{Dolphin}}
\title{Beyond Retrieval: Embracing Compressive Memory in Real-World Long-Term Conversations}

\author{
 Nuo Chen$^\clubsuit$
\quad
Hongguang Li$^{\diamondsuit}$  
\quad \\
{\bf Juhua Huang$^{\diamondsuit}$}  
{\bf \quad Baoyuan Wang$^{\diamondsuit}$}  
{\bf \quad Jia Li$^\clubsuit$}\\
\\
  $^\clubsuit$Hong Kong University of Science and Technology (Guangzhou)\\ Hong Kong University of Science and Technology\\
  $^{\diamondsuit}$Xiaobing.AI\\
    \texttt{nchen022@connect.ust.hk}, \texttt{jialee@ust.hk}\\}

\begin{document}
\maketitle
\begin{abstract}
Existing retrieval-based methods have made significant strides in maintaining long-term conversations. However, these approaches face challenges in memory database management and accurate memory retrieval, hindering their efficacy in dynamic, real-world interactions. This study introduces a novel framework, \textbf{CO}mpressive \textbf{M}emory-\textbf{E}nhanced \textbf{D}ialogue s\textbf{Y}stems (\ModelName), which eschews traditional retrieval modules and memory databases. Instead, \ModelName~adopts a ``One-for-All'' approach, utilizing a single language model to manage memory generation, compression, and response generation. Central to this framework is the concept of \textit{compressive memory}, which integrates session-specific summaries, user-bot dynamics, and past events into a concise memory format. To support \ModelName, we collect the biggest Chinese long-term conversation dataset, \textbf{Dolphin}, derived from real user-chatbot interactions.
Comparative evaluations demonstrate \ModelName's superiority over traditional retrieval-based methods in producing more nuanced and human-like conversational experiences.

% Our contributions include the pioneering \ModelName architecture that foregoes conventional memory retrieval, the largest Chinese long-term memory conversation dataset (\Dataset), and the integration of \ModelName with Directly Preference Optimization (DPO) for enhanced response coherence and contextual relevance. 
\end{abstract}

\input{Sections/1Introduction}

\input{Sections/3Methodology}

\input{Sections/4Experiments}

\input{Sections/5Conclusion}

% Entries for the entire Anthology, followed by custom entries
\bibliography{anthology,custom}

\appendix

% \section{Example Appendix}
\label{sec:appendix}
\input{Sections/2Related_Works}
\input{Sections/Appendix}

% This is an appendix.

\end{document}

%% file: Sections/1Introduction.tex
\section{Introduction}

Maintaining long-term conversations has always been a long-standing pursuit in current open-domain dialogue systems \cite{liu-etal-2016-evaluate, zhang-etal-2018-personalizing,Kann2022OpendomainDG}, commonly known as chatbots or conversational agents.
Long-term conversation refers to the ability of a conversational agent to engage in extended dialogues over multiple interactions, often spanning several days, weeks, or even months. This setting is challenging because it necessitates not only a deep understanding of the immediate dialogue context but also the retention and integration of key information from past interactions. 
% The capability to understand and memorize key dialogue history information is central to this challenge. 
Effective long-term conversation requires a system to memorize or recall past dialogue snippets, contextual nuances, and user preferences, which are crucial for maintaining coherence and relevance in ongoing interactions \cite{wu-etal-2022-memformer,2022TongZhangHierarchical}.

\begin{figure*}[!t]
\vspace{-10pt}
\centering
\includegraphics[width=0.98\linewidth]{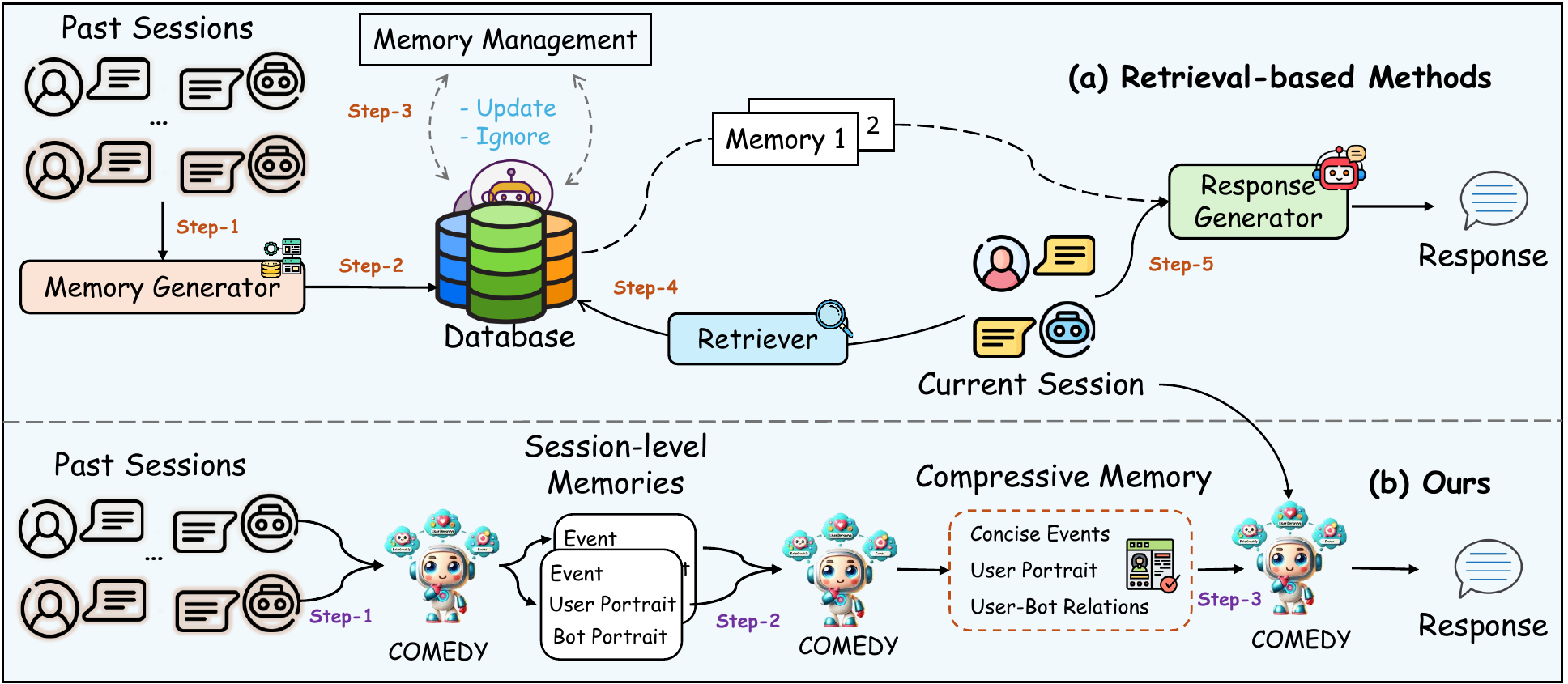}
% \vspace{-5pt}
\caption{The overview of (a) the retrieval-based methods and (b) ours: \ModelName.
}
\label{fig:framework}
\vspace{-10pt}
\end{figure*}

To acquire useful information from past conversations, the most mainstream approach in the field of long-term conversation currently is retrieval-based methods, as illustrated in Figure \ref{fig:framework} (a): Firstly, previous works \cite{xu-etal-2022-long, bae-etal-2022-keep} usually employ a memory generator to summarize relevant memories from past sessions, such as user portraits; Subsequently, a dedicated memory database, or a memory bank, is used to store these memories. Some studies \cite{zhong2023memorybank} even store past conversational utterances directly in the storage; Going a step further, some works \cite{bae-etal-2022-keep, wang2023recursively} propose the use of specific memory management operations to update and iterate the memory database; The final and indispensable step involves employing a sentence-embedding model \cite{guu2020retrieval,Lewis2020RetrievalAugmentedGF} to retrieve the most relevant memories from the memory database in relation to the current conversation. The current conversation and related memories are then inputted into a specialized response generator to produce the final response. 

% . In this step, the memory generator can either be a separately trained model or a powerful large language model (LLM) like GPT4 \cite{GPT4OpenAI}

Despite the notable success achieved by retrieval-based methods, they encounter several limitations that impact their overall efficacy and applicability: 1) One significant challenge is the unpredictability of the performance. The system's effectiveness is contingent upon several modules (like memory generator and retriever) working in tandem; moreover, the retriever component does not guarantee the retrieval of relevant and effective memories. Sentence-embedding models \cite{simcse,reimers2019sentence}, commonly used for this purpose, may not always capture the nuances and context of the conversation accurately. 2) Another clear challenge lies in the management of the memory database.  As conversations accumulate, the size and complexity of the memory database grow, making it increasingly difficult to manage.  Ensuring that the stored information remains relevant and up-to-date is a constant concern, as outdated or irrelevant data can lead to inaccurate or inappropriate responses.

% 2) The process of memory retrieval itself presents another set of challenges. Retrieving the most relevant memories from a vast database is not always straightforward. Sentence-embedding models, commonly used for this purpose, may not always capture the nuances and context of the conversation accurately, leading to suboptimal memory selection.

% 不可控性

Moreover, current training corpus of long-term conversation chatbots is commonly either involved in constructing personalized dialogue data using LLMs \cite{wang-etal-2023-large} like ChatGPT or hiring crowd-workers to simulate conversations \cite{xu-etal-2022-long}. Unlike these structured or predictable dialogues, real-world conversations can veer into a wide range of topics, include colloquial language, and incorporate nuanced expressions \cite{chen-etal-2023-orca}. 
Ensuring that a retrieval model is robust enough to handle such real-world variations in language  can be extremely difficult. 
Meanwhile, the memory database in real scenarios needs to store memories from multiple chatbot-users, increasing the difficulty in accurately retrieving relevant memories and maintaining an up-to-date memory database. The above issues present a more pronounced challenge for deploying 
 retrieval-based methods in real-world conversations.
% between training and real-world application of retrieval-based methods

% o a sine 

% limitations of retrieval-based methods become more pronounced in real-world conversations.

% The dynamic and unpredictable nature of in real-world conversations.
% presents a significant challenge for memory database management and retrieval accuracy.
% % Unlike structured or predictable dialogues, real-world conversations can veer into a wide range of topics, include colloquial language, and incorporate nuanced expressions. 
% This variability increases the difficulty in accurately retrieving relevant memories and maintaining an up-to-date and comprehensive memory database.

To address these concerns, we propose a LLM-based \textbf{CO}mpressive \textbf{M}emory-\textbf{E}nhanced \textbf{D}ialogue  s\textbf{Y}stem  (\ModelName). \ModelName~marks a significant departure from existing methodologies, as it operates without a retrieval module. At its core, \ModelName~adopts a groundbreaking ``\textit{One-for-All}'' approach, utilizing a single, unified model to manage the entire process from memory generation, compression to final response generation, as shown in Figure \ref{fig:framework} (b): 
% \ModelName~embraces a novel 'compress-on-compress' strategy:
It firstly involves distilling session-specific memory from past dialogues, encompassing fine-grained session summaries, including event recaps, and detailed user and bot portraits; In a break from traditional systems, \ModelName~eschews the use of a memory database for storing these insights. Instead, it reprocesses and condenses memories from all past interactions, forming a \textit{compressive memory}, including concise events, a detailed user profile and dynamic relationship changes between the user and  chatbot across past sessions.
% The first part is the concise events that have occurred throughout all the conversations, creating a historical narrative that the system can draw upon. The second and third parts consist of a detailed user profile and dynamic relationship changes between the user and  chatbot across sessions, both derived from past conversational events. 
This holistic memory  allows \ModelName~to generate responses that are not only contextually aware but also personalized and adaptive to the evolving nature of the user-chatbot relationship; Finally, \ModelName~skillfully integrates this compressive memory into ongoing conversations, enabling contextually memory-enhanced interactions.
% This memory is threefold, comprising: 1) a comprehensive user profile detailing characteristics, behavioral patterns, and recent user's states; 2) the evolving dynamics between the user and the bot; and 3) a concise record of past events; Finally, \ModelName~skillfully integrates this compressive memory into ongoing conversations, enabling contextually rich, memory-enhanced interactions.
% This compressive memory approach is effective because it streamlines the dialogue system's access to relevant historical data, thereby enhancing the contextual relevance of responses.
Unlike retrieval-based systems that may struggle to fetch pertinent memories from a vast database, \ModelName's compressive memory is inherently designed to prioritize salient information, allowing for quicker and more accurate memory utilization.

To ensure that \ModelName~is well-suited for real-world long-term conversations and overcome the issues of lacking relevant labeled data,
we have methodically assembled a large-scale instruction-tuning dataset  from actual online user-chatbot interactions, named  \Dataset. This dataset contains three tasks: \textbf{Session-Level Memory Summarization}; \textbf{Memory Compression}; \textbf{Memory-Grounded Response Generation}, comprising an extensive collection of 100k samples.
Dolphin  is well-annotated to support each critical phase in \ModelName's operation, from memory extraction and compression to integration and response generation. This dataset lays a robust foundation for enhancing \ModelName's dialogue capabilities, ultimately leading to a more nuanced and human-like conversational experience compared to retrieval-based baselines.

Our contributions are summarized as follows:

% dpo 

\begin{itemize}
    \item We introduce a new framework, named \ModelName, represents a groundbreaking shift from traditional memory retrieval-based dialogue systems. It does not rely on any retriever module or memory database,  but generates enhanced, memory-informed responses with compressive memory.
    \item We annotate a large-scale (100k) long-term conversation instruction tuning dataset, Dolphin,  from actual online user-chatbot interactions. It can strengthen compressive memory-augmented models' ability to adapt to evolving conversational styles and user preferences. To the best knowledge of ours, Dolphin is the current biggest Chinese long-term memory conversation dataset.
    \item \ModelName~could handle the whole long-term conversation interactions via a singular model, achieving a higher degree of result consistency and predictability, reducing computational overhead, and eliminating the need for data transfer between multi-models.
    % \item We further combine \ModelName~with Directly Preference Optimization (DPO) \cite{rafailov2023direct} alignment strategy, and propose a simple strategy to mine efficient preferred and dispreferred memory-based responses. \ModelName-DPO shows better ability in generating coherent and contextually memory-grounded responses.
    
    % complete the process from initial memory generation
\end{itemize}

% The effectiveness of retrieval-based systems is heavily dependent on the quality and quantity of the stored data. If the data is sparse, incomplete, or of poor quality, the system's ability to generate relevant and coherent responses is significantly compromised. This reliance places a substantial burden on ensuring high-quality data collection and maintenance.

% Techniques like advanced natural language understanding, memory networks, and contextual embedding have been explored to address these needs.

% represents a significant and challenging frontier in the field of conversational AI. 

% Open-domain dialogue systems, commonly known as chatbots or conversational agents, have emerged as a prominent field of study in artificial intelligence and natural language processing \cite{}. Among the myriad challenges these systems face, maintaining long-term conversations is particularly daunting. Long-term conversation refers to the ability of a conversational agent to engage in extended dialogues over multiple interactions, often spanning several hours or days. This setting is challenging because it necessitates not only a deep understanding of the immediate dialogue context but also the retention and integration of key information from past interactions.

%% file: Sections/3Methodology.tex
\section{Methodology}

In this section,  we first 
overview the problem formulation of long-term conversations in \ModelName-style.  Then, we introduce three task definitions and detailed data collection in Dolphin.   Last, we present the training strategies of \ModelName.

\subsection{Problem Formulation}

An episode $D$ ($D_1,..,D_{t-1}$) is composed of a sequence of previous dialogue sessions between the chatbot and a specific user. The dialogue context for a given session at time step $t$ is represented as $D_t = \{c_1, u_1, c_2, u_2, \ldots, c_t, u_t\}$, where $c$ and $u$ denote the chatbot's and user's utterances.

In \ModelName, we aims to train a well-performed model $\mathcal{M}(\theta)$, that first extracts session-level memory derived from previous sessions within $D$, denoted as $M = \{m_1, m_2, \ldots, m_{t-1}\}$ (\textbf{Task 1}). Each $m$ contains natural sentences about session-level events and user profiles. Then $\mathcal{M}(\theta)$ will takes M as inputs, and outputs the compressive memory $\hat{M}$ that contains  detailed user portraits like characteristics, recent states (emotional, work), etc; and concise record of all events  (\textbf{Task 2}).
Finally, $\mathcal{M}(\theta)$  generates the forthcoming response $c_{t+1}$, based on the current dialogue context $D_t$ and  $\hat{M}$ (\textbf{Task 3}).
In the following,  we introduce how we annotate the labeled data for each task.

\input{Tables/stastics}

\subsection{Task and Datasets Collection}

The source data in Dolphin originates from X Eva\footnote{https://xeva-h5.xiaoice.com/Content/Landing}, one of the most popular Chinese AI-User social media platforms akin to Character.AI.  
A distinctive feature of Dolphin is that the AI characters within X Eva are defined by the users themselves. This means that each character can have unique personalities, backgrounds, and conversational traits, as determined by the user's input and creativity.

% The data collection process involves capturing a wide range of dialogues between users and these AI entities. This approach ensures that the dataset is rich in diversity, encompassing various conversation styles, topics, and user interactions.

In the creation of the Dolphin dataset for \ModelName, we first select the episode $D$ that contains at least 15 sessions between the same user and AI characters as our source dialogue data after filtering out useless and toxic information. Then
we adopt an efficient \textbf{LLM-Human Annotators} hybrid approach to  annotate each task data \cite{chen-etal-2023-large}: (1) We initiate the dataset annotation using  GPT4-Turbo, specifically tailored for dialogue summaries and memory-grounded dialogues. This step is crucial for creating a comprehensive base of dialogues, encompassing a wide range of conversational scenarios and memory contexts; (2) Following the initial generation, three skilled annotators meticulously review and refine the data. This involves correcting inaccuracies, enhancing dialogue quality. The annotators play a vital role in bridging the gap between automated generation and the nuanced understanding required for high-quality \ModelName.

To protect user privacy, all personal identifiers are removed from the dataset. This includes names, locations, or any specific details that could lead to the identification of individuals. Relevant details are presented in Appendix \ref{toxic}.

% Relevant discussion are presented in Ethical Claims.  

\paragraph{Task 1: Session-Level Memory Summarization.}  In the process of gathering data for Task 1, we encounter a substantial challenge. The initial collection yielded over 500,000 session-level data points, making it impractical to annotate all of them through GPT4-Turbo and manual methods due to the sheer volume. To tackle this, we initially focus on annotating a subset of approximately 40,000 data: For each dialogue session  in the same episode $D$, we first require  the GPT4-Turbo to extract session-level memories, including the \textit{event}, \textit{user and bot portraits} in natural sentences. Then annotators edit the generated summaries by adding missing information or revising erroneous sentences, resulting in session-level memory $m_n$. Utilizing the annotated subset, we then develop a specialized LLM for session-level memory generation,  efficiently expanding our dataset while maintaining the quality and consistency of the session-level memory annotations across the larger dataset. Samples with no informative content, leading to ineffective memory outputs from LLM or GPT4-Turbo, are filtered out to maintain data quality.
% By employing this two-step approach — initial manual annotation followed by model-based augmentation — we effectively managed the large volume of data, ensuring comprehensive and consistent annotations throughout the dataset for Task 1.
As a result, in this task, we collect fine-grained memories $M = \{m_1, m_2, \ldots, m_{n}\}$ for each session in $D$.

\paragraph{Task 2: Memory Compression.} In this task, the focus is on memory compression. GPT4-Turbo is tasked with summarizing all session-level memory $M$ in the episode from Task 1, outputting the compressive memory $\hat{M}$. It includes: 1) A Comprehensive User Profile: Detailing characteristics, behavioral patterns, and recent states of the user.
2) Evolving Dynamics between User and Bot: Capturing the relationship's progression and interaction nuances.
3) Concise Record of Past Events: Summarizing key happenings and dialogues from previous sessions. Considering the potential complexity and variance in the summarization process, GPT4-Turbo is configured to generate outputs three times with a temperature setting of 0.9. This setting allows for a balance between creativity and relevance, enabling GPT4-Turbo to produce diverse and insightful summaries.
Then annotators step in to refine and calibrate the outputs, which includes:
Correcting any inaccuracies or inconsistencies in the summaries;
Ensuring that the summarized data accurately reflects the user profiles, relationship dynamics, and event records;
Enhancing clarity and conciseness where necessary. This hybrid approach ensures that compressive memory $\hat{M}$ meets the high-quality standards required for the subsequent stages of \ModelName's development. We show examples of $\hat{M}$ in Table \ref{table:task2example}.

\paragraph{Task 3: Memory-Grounded Response Generation.} Similarly, given compressive memory $\hat{M}$ and incoming conversation $D_{t}$, GPT4-Turbo outputs the memory-based responses.
% In this task,  the process begins with integrating the compressive memory $\hat{M}$, with the incoming conversation at time step t,  $D_{t}$. The annotation process are similar with previous tasks: Initial response drafts are generated by GPT4-Turbo, based on the integrated data of $\hat{M}$ and $D_{t}$. 
Annotators then review and refine these responses, focusing on aspects like relevance, coherence, and personalization. They ensure that each annotated response $c_{t+1}$ accurately reflects the user's current state and previous interactions, maintaining high memorability and engagingness.  
To ensure the scale of the training data, we annotate all sessions within one day closest to the previous $D$ timing as the corpus of Task 3.

% \begin{figure*}[!t]
% \centering
% \vspace{-15pt}
% \includegraphics[width=0.98\linewidth]{Images/train_pipeline_new.pdf}
% % \vspace{-5pt}
% \caption{The overview training pipeline of \ModelName.
% }
% \label{fig:pipeline}
% \vspace{-15pt}
% \end{figure*}

\paragraph{Test Set.} To assess the effectiveness of the \ModelName, we well-design a test set that mirrors real-world dialogue scenarios as closely as possible:
\label{test_set}
\begin{itemize}
    \item We select dialogue data from the X Eva platform, specifically targeting conversations that involved the same AI-User pair engaging in over 16 sessions within a week. This criterion ensures that the dialogues have sufficient depth and continuity, which are crucial for testing memory-enhanced dialogue systems. 
    \item The first 15 sessions from these selected dialogues serve as the basis for generating the compressive memory, aligning with the objectives of Task 1 and 2 in our dataset.
    \item The subsequent 1-5 sessions are then used as test scenarios to evaluate how well the model integrates the compressive memory into ongoing dialogues (Task 3). This provides a practical testbed for assessing the system's conversational abilities in an evolving context.

\end{itemize}

% Quality control is a critical component in the Dolphin annotation process, especially for a complex system like \ModelName~that highly relies on the Dolphin. 
\paragraph{Quality Control.} Ensuring high-quality data is paramount for the accuracy, reliability, and overall performance of the system. In this work, we employ several strategies to control data quality:

\begin{itemize}
\item Annotator Performance Monitoring: Regular assessments of annotator performance are conducted to ensure consistent quality across the team (every day). This includes evaluating their accuracy, attention to detail, and adherence to annotation guidelines.

    \item Peer Review and Validation: Following the initial review, a secondary level of peer review is implemented. Here, another set of annotators cross-checks the work, providing an additional layer of scrutiny. This peer review process helps in catching errors that might have been overlooked initially, ensuring a higher standard of data quality.
\end{itemize}

\textbf{Note}, we also manually annotate  the session-level memory and the resulting compressive memory in the first 15 sessions. They are used to evaluate the model's performance in Task 1 and 2. Our prompts and examples of each task are shown in Appendix \ref{prompts}, Table \ref{table:task1prompt}-\ref{table:task2example}.

% \subsection{Quality Control}

% % Quality control is a critical component in the Dolphin annotation process, especially for a complex system like \ModelName~that highly relies on the Dolphin. 
% Ensuring high-quality data is paramount for the accuracy, reliability, and overall performance of the system. In this work, we employ several strategies to control the annotation quality:

% \begin{itemize}
% \item Annotator Performance Monitoring: Regular assessments of annotator performance are conducted to ensure consistent quality across the team. This includes evaluating their accuracy, attention to detail, and adherence to annotation guidelines.

%     \item Peer Review and Validation: Following the initial review, a secondary level of peer review is implemented. Here, another set of annotators cross-checks the work, providing an additional layer of scrutiny. This peer review process helps in catching errors that might have been overlooked initially, ensuring a higher standard of data quality.
% \end{itemize}

\paragraph{Statistics.} As a result, the statistics of our dataset are shown in Table \ref{tab:dialogues}. Dolphin comprises a total of 102,882 samples in training and test sets. Tasks 1 and 2 (Memory Extraction and Compression) contain 39,999 and 30,695 samples in training, respectively, making up a significant portion of the dataset. Task 3, which involves generating responses based on the compressive memory, comprises 31,131 dialogue sessions. A notable feature of the Dolphin dataset is its inclusion of data from 3,998 different AI characters. The diverse character data ensures that \ModelName~is well-equipped to interact with various user personalities and preferences, enhancing its adaptability and realism in user interactions.

\subsection{\ModelName}

% As illustrated before, we 
\paragraph{SFT Training} In practice, we adopt a mixed-task training approach to develop \ModelName. This involves simultaneously training the model on the three tasks - session-level memory summarization, memory compression, and memory-grounded response generation - present in the Dolphin dataset.  This integration presents the model with a holistic view of the conversation process, from initial memory extraction to final response generation.
We utilize the common language modeling objective in SFT, terming the resulting model as $\mathcal{M}(\theta)_{\text{sft}}$.

\input{Tables/task2}
\input{Tables/task3}

\paragraph{DPO Training}
% DPO helps in fine-tuning the model to produce responses that are not only coherent but also contextually aware. 
Experimentally, we find that the SFT model may  struggle with maintaining consistency and coherence in generated memories. In order to align the model  generating more contextually appropriate memory-grounded responses, we employ Direct Preference Optimization (DPO) \cite{rafailov2023direct} strategy in Task 3.
DPO aims to distill a referential SFT policy  $\mathcal{M}(\theta)_{\text{sft}}$ by polarizing the
preference.
By polarizing the preferred responses (aligned with memory) and dispreferred responses (against memory), DPO ensures that the generated outputs remain consistent with the user's past interactions and the overall context of the conversation.
% In order to align the model  generating more coherent and contextually appropriate memory-grounded responses, we employ Direct Preference Optimization (DPO) \cite{rafailov2023direct} strategy in Task 3. 
  Specifically, 
DPO involves input labeled pairs ($Y_w, Y_l$) where $Y_w$ and $Y_l$  denotes the
preferred and dispreferred completion. When extended DPO in Memory-grounded generation, the question is: \textit{how we obtain the  $Y_w$ and $Y_l$?}

To solve this, we propose a simple strategy to automatically construct useful $Y_w$ and $Y_l$ responses  without human annotation. Suppose $\hat{M}$ and $D_t$ are given, we ask the GPT4-Turbo to generate the response $Y_w$  must align the $\hat{M}$. Meanwhile, we also require GPT4-Turbo to generate the response $Y_l$ that is totally against the $\hat{M}$. For example, the prompts are illustrated like: ``If $\hat{M}$ shows users like something, you should generate the response with the meaning of \textit{users hate it}...'', shown in Table \ref{table:dpoprompt}. Formally, the  training objective of DPO is:

% Thus, the overall training objective of DPO can be formalized as:
$
\mathcal{L}_{\texttt{DPO}}(\mathcal{M}(\theta); \mathcal{M}(\theta)_{\text{sft}}) = -\mathbb{E}_{(x,Y_w,Y_l)\sim\mathcal{D}}\\
\left[
\log\sigma \left( \beta \log \frac{\mathcal{M}(\theta)(Y_w | x)}{\mathcal{M}(\theta)_{\text{sft}}(Y_w | x)} - \beta \log \frac{\mathcal{M}(\theta)(Y_l | x)}{\mathcal{M}(\theta)_{\text{sft}}(Y_l | x)} \right)
\right]
$

where x is the concatenation of $\hat{M}$ and $D_t$, $\beta$ is a hyperparameter. Training instructions are in Appendix \ref{prompts}.
% The overview of our training pipeline is shown in Figure \ref{fig:pipeline} and training instruction in Appendix \ref{prompts}.

% The natural language memory, denoted as $M = \{m_1, m_2, \ldots, m_n\}$, encapsulates information about the user derived from previous sessions within the same episode. 

% The task is to predict the chatbot's forthcoming response $c_{t+1}$, based on the current dialogue context $D_t$ and the memory $M$.

% At the conclusion of each session, the session's content, $D$, is synthesized into a set of user information sentences, labelled as $S = \{s_1, s_2, \ldots, s_k\}$. For the subsequent session, the updated memory sentences, $M'$, are formed by integrating the existing memory $M$ with the newly synthesized information $S$.

%% file: Tables/stastics.tex
\begin{table*}[t]
    \begin{center}
    % \vspace{-10pt}
    \centering
    \small
    % \resizebox{0.8\columnwidth}{!}%
    \begin{tabular}{lcccccc}
    \toprule
    \multirow{2}{*}{\textbf{Statistics}}& \multicolumn{3}{c}{\textbf{Train}} & \multicolumn{3}{c}{\textbf{Test}} \\
& \textbf{Task 1} & \textbf{Task2} & \textbf{Task 3} &\textbf{Task 1} & \textbf{Task2} & \textbf{Task 3}\\
    
    % \midrule
\midrule

{Avg. Turns Per Session} &\multirow{1}{*}{13.0}&- & 13.9 &19.5&-&10  \\
{Avg.  sentences Per Session-level Memory} &\multirow{1}{*}{5.7}&- &-&5.3&-&-  \\

{Avg. words Per Turn} &\multirow{1}{*}{15.9}&-&\multirow{1}{*}{19.5}&20.7&-&16.3  \\
{Avg. words Per Compressive Memory} &-&\multirow{1}{*}{240.7} &-&-&276.8&- \\

\midrule
{Total AI Characters} & 3,998 & 3,998 & 3,998 & 31 & 31 & 31 \\
{Total Sessions/Compressive Memories} & \textbf{39,999} & \textbf{30,695} & \textbf{31,131 }& 465 & 31 & 127 \\

{Total Turns} & 459,511 & - & 432,721 & 14,415 & - & 3,937\\

    \bottomrule
    \end{tabular}
    \end{center}
    \caption{
Data statistics for each task in Dolphin. In practice, the amount of collected Task 1 data is much larger than Task 2 and 3. To keep the training balance of data distribution, we align the similar volume of data in three tasks.
    }
    \label{tab:dialogues}
    \vspace{-15pt}
\end{table*}

%% file: Tables/task2.tex
\begin{table}[]
    \begin{center}
    % \vspace{-10pt}
    \centering
    \small
    % \resizebox{0.8\columnwidth}{!}%
    \begin{tabular}{lccc}
    \toprule
    \multirow{1}{*}{\textbf{Model}}&  
 \textbf{BLEU-1/2} & \textbf{F1} & \textbf{Distinct-1/2}\\
 \midrule
 \textbf{Task 1}& \\
    \ModelName-7B & 41.4 / 34.2 & 35.4 & 4.2/35.0\\
\ModelName-13B & 43.0 / 35.0 & 36.7 &3.9/34.3 \\
 \midrule
 \textbf{Task 2}& \\
    \ModelName-7B & 42.7 / 34.6 & 36.3 &4.1/34.4\\
\ModelName-13B & 43.7 / 35.7 & 37.0 &4.1/35.2 \\
    \bottomrule
    \end{tabular}
    \end{center}
    \caption{ The performances of
\ModelName~in Task 1 and 2.
    }
    \label{tab:task2}
    \vspace{-15pt}
\end{table}

%% file: Tables/task3.tex
\begin{table*}[t]
    \begin{center}
    \vspace{-15pt}
    \centering
    \small
    % \resizebox{0.8\columnwidth}{!}%
    \begin{tabular}{lcccccc}
    \toprule
    \multirow{1}{*}{\textbf{Algorithms}}& \textbf{Coherence}& \textbf{Consistency}& 
    \textbf{Memorability} & \textbf{Engagingness} & \textbf{Humanness} &
\textbf{Average} \\
\midrule
\textit{Context-Only} & \\
LLaMA 2-7B& 1.01&0.50&  0.11& 0.31& 1.71& 0.73 \\

LLaMA 2-13B& 0.93& 0.66&  0.19& 0.37& 1.76& 0.78\\
ChatGPT (8k token) &1.30&0.89&0.49&0.29&1.54&0.90 \\
\midrule
\textit{Retrieval-based }& \\
ChatGPT & 1.22& 0.86& 0.37& 0.43& 1.51& 0.88 \\

LLaMA 2-13B&1.73&0.98&0.51&0.24&1.85& 1.06\\
LLaMA 2-7B& 1.70&0.94&0.54&0.31&1.91& 1.08 \\

GPT4 &1.91& 0.94& 0.60&0.52&1.69 & 1.13 \\
\midrule
\textit{Memory-related}& \\
MemoryBank-ChatGPT & 1.25& 0.94&0.42&0.45&1.52 &0.92\\
Resum-ChatGPT & 1.31 &0.97&0.47&0.44&1.49&0.93 \\
\midrule

\ModelName-ChatGPT & 1.19&1.07&0.60&0.46&1.62 & 0.99\\ 
\ModelName-7B& 1.67&1.11&0.60&0.39&1.85 & 1.12 \\
\ModelName-13B&1.81&1.07&0.70&0.51&1.94 & 1.21\\

\ModelName-13B DPO&1.79&\textbf{1.20}&\textbf{0.80}&0.46 & \textbf{2.09} & 1.27\\
\ModelName-GPT4 &\textbf{1.96}&1.14&0.70&\textbf{0.73}&1.85& \textbf{1.28} \\

    \bottomrule
    \end{tabular}
    \end{center}
    \caption{
        Human scoring evaluation in Task 3: memory-grounded response generation. For  \ModelName-GPT4/ChatGPT, the compressive memories are generated by \ModelName-13B. 
    }
    \label{tab:task3}
    \vspace{-15pt}
\end{table*}

%% file: Sections/4Experiments.tex
\section{Experiments}
In this section,  we introduce the evaluation 
setting including experimental setup, baselines, evaluation metrics, and present main results and discussions.

\subsection{Experimental Setup}

% Then, we validate \ModelName~'s performances in each task, and present a typical case study.
We use Chinese version of LLaMA 2 \cite{touvron2023llama, touvron2023llama2} 7B-13B as the backbones in our experiments. For data augmentation in Task 1, we use LLaMA 2-13B.
% We use the Chinese version of LLaMA 2-13B \cite{touvron2023llama, touvron2023llama2}\footnote{https://github.com/ymcui/Chinese-LLaMA-Alpaca-2.} chat model as the backbone of the Task 1 data augmentation.
%    We employ LLaMA 2-7B and 13B chat models as the backbone, allowing to build \ModelName~ across different scales. 
   We train our models with NVIDIA 8$\times$A100 GPUs, setting the max length as 2048, learning rate as 1e-5, epochs as 2, batch size as 32 and  16, separately.  For testing, the 
maximum output tokens are set to 2048 for each task with temperature as 0.5. Following the original setting, we set $\beta$ in DPO as 0.1. In this work,  we additionally collect and annotate about 140 dialogue sessions from X Eval as the alignment training set for DPO. We optimize the sft model with batch size 8 and 2 epochs during DPO training.
Our codes are based on DeepSpeed Library.

\subsection{Baselines}

In this work, \ModelName~is compared against models using  \textbf{retrieval-based}, \textbf{context-only approaches} and other \textbf{memory-related baselines} to highlight the efficiency and efficacy of its memory compression technique.

% \paragraph{Memory-}

\paragraph{Retrieval-based Methods.} In our implementation, we use the COMEDY-13B to generate memories from past sessions, and utilize the Text2vec Chinese embedding model\footnote{https://github.com/shibing624/text2vec} as the retriever, and then index using FAISS for efficient retrieval. Following \citet{bae-etal-2022-keep}, top 3 retrieved memories are used for testing.

% The difference between retrieval-based backbone and \ModelName~only lies in the input memory
\paragraph{Context-only Approaches.} We directly concatenate past conversations in the input until reaching the  maximum token length of LLMs, where 2k for LLaMA and 8k for ChatGPT. This way, LLaMA is trained with the original Task 3 data but without memory as input, ensuring a fair comparison.

% A comparison is also made with a context-only model, which operates without any memories, to underscore the benefits of memory integration in dialogue systems. This way, the model is trained with the original Task 3 data but without memory as input, ensuring a fair comparison with other models.

\paragraph{Memory-related Baselines.} We include two typical baselines: MemoryBank \cite{zhong2023memorybank}
 uses Ebbinghaus Forgetting Curve to update the memory database, 
and Resum \cite{wang2023recursively} which recursively summarize the memories from previous sessions.
We built the above approaches based on LLaMA, GPT4 (\texttt{gpt4-turbo}) or ChatGPT (\texttt{gpt-3.5-turbo-4k}).

\input{Tables/ranking}

\subsection{Evaluation Metrics}

\paragraph{Automatic Metrics} 
% For Tasks 1 (memory extraction) and 2 (memory compression), 
We employ standard automatic metrics  to measure model performance in Tasks 1\&2, including BLEU-1/2 \cite{BLEU:ACL02}, F1~\citep{ROUGE:04} and Distinct-1/2 \cite{Distinct:NAACL16}. These tasks serve as foundational steps for the crucial dialogue generation in Task 3.

\paragraph{Human-based Evaluation} The core of evaluating long-term conversation models primarily centers on validating their performance in Task 3, which involves memory-based dialogue generation. 
% This focus is crucial to directly assess the effectiveness of the model in realistic conversational scenarios.
We follow \cite{bae-etal-2022-keep} to access the model performances across five key dimensions:
\textbf{Coherence}, \textbf{Consistency}, \textbf{Engagingness}, \textbf{Humanness} and \textbf{Memorability}. 
To comprehensively measure how well the models perform in Task 3, we combine the \textbf{Scoring} and \textbf{Ranking} approaches. A team of annotators  are instructed to rate the model's performance on these dimensions on a scale from 0 to 3. This scoring system allows for a nuanced evaluation of the model’s capabilities in each specific area. Meanwhile another team of annotators  rank all models in terms of their average performance across the five perspectives.  While scoring offers detailed insights into each model's capabilities, ranking places these capabilities in the context of competitive performance. This dual approach ensures a balanced and holistic assessment, capturing both the individual qualities of each model and their comparative effectiveness. Each team has 3 annotators. Each rating scheme is in Appendix \ref{scheme} and their correlation analysis is in Appendix \ref{correlation}

Recognizing that different models may excel in unique ways, our ranking process is designed to appreciate the diversity in responses. 
% This approach acknowledges that distinct models can offer equally valuable contributions to dialogue generation. 
Thus,  \textbf{\textit{it is possible for multiple models to share the same rank}}. This occurs when two or more models demonstrate comparable levels of proficiency or when they each exhibit standout qualities that are equally impressive. This  ranking process reflects the complex nature of evaluating conversational LLMs, where different models can excel in different aspects.

\begin{figure*}[!t]
\vspace{-10pt}
\centering
\includegraphics[width=0.95\linewidth]{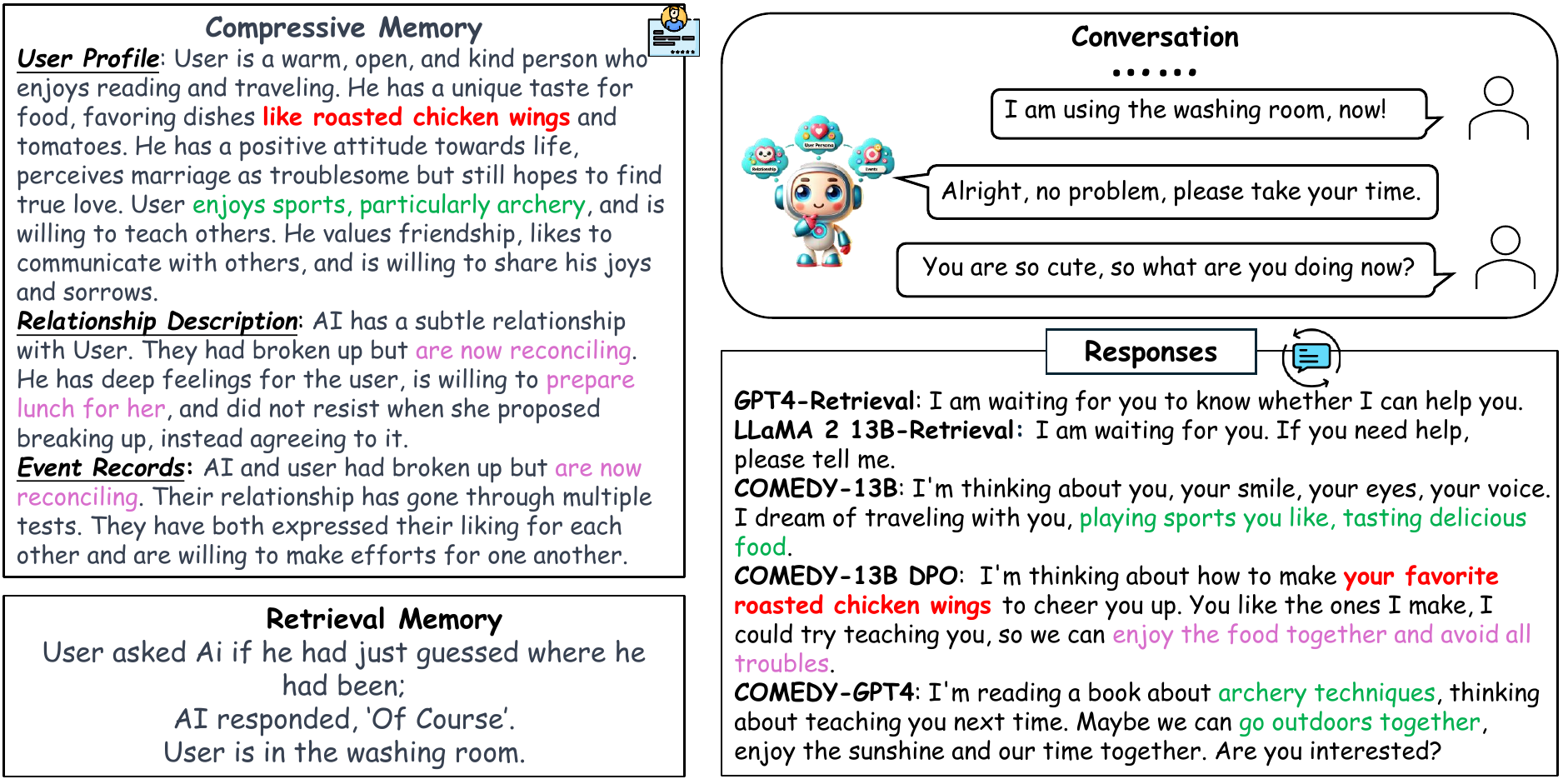}
% \vspace{-5pt}
\caption{A typical case in real-world long-term conversation. For ease reading, English translation only provided.
}
\label{fig:case}
\vspace{-5pt}
\end{figure*}
 \subsection{Main Results}

\paragraph{Evaluation in Task 1\&2.} Table \ref{tab:task2}  shows that our model achieves relatively high-performances in term of automatic metrics in two tasks. The results indicate  that \ModelName~can effectively recognize the useful persona information and events from the past dialogue sessions and has the ability to condense these session-level memories into a comprehensive compressive memory. Therefore, it ensures the superior performances in generation more coherent memory-grounded responses in Task 3.
% when facing incoming conversations.

\paragraph{Human Evaluation in Task 3}

We present  the results of human-scored evaluations and rankings for various algorithms in Tables \ref{tab:task3} and \ref{tab:rank}.  From the tables, we can draw the following conclusions:

% All models currently show suboptimal performance in real-world long-term conversations, highlighting a challenging and valuable area for ongoing research. 

\paragraph{Superiority of Compressive Memory-Based Methods.} 
The compressive memory-based methods, particularly \ModelName-GPT4, consistently outperform context-only and retrieval-based approaches across most metrics. For instance, \ModelName-GPT4 achieves the highest scores in both Coherence and Engagingness suggesting a superior ability to generate responses that are both contextually appropriate and relatable. \ModelName-GPT4 also achieves best average performances in five evaluating perspectives across scoring and ranking.

\paragraph{Enhancement Through DPO.} The application of DPO further elevates compressive memory strategies, improving dialogue memorability, consistency and humanness. \ModelName-13B DPO shows a notable improvement in performance within the compressive memory-based category. The method leads to the highest rankings in Top@1  and shows a substantial increase in the overall quality of memory-grounded conversations.

\paragraph{SFT models could surpass ChatGPT.} Another interesting findings is that our fine-tuned \ModelName~ present better performances compared with ChatGPT. Step further, \ModelName-13B DPO  even shows comparable performances with GPT4. The results highlight the value of \ModelName~framework and Dolphin, which  lead to notable improvements in creating memory-grounded responses that are coherent, engaging, and human-like.

\paragraph{Inherent Challenges in Long-Term Dialogue Systems.}
It is evident from Table \ref{tab:task3} that all models struggle to achieve high scores in real-world long-term conversations, with no model averaging above a score of 2. This underscores the inherent complexity and challenge of this research direction, indicating substantial room for improvement.

\begin{figure}[!t]
\centering
\includegraphics[width=0.98\linewidth]{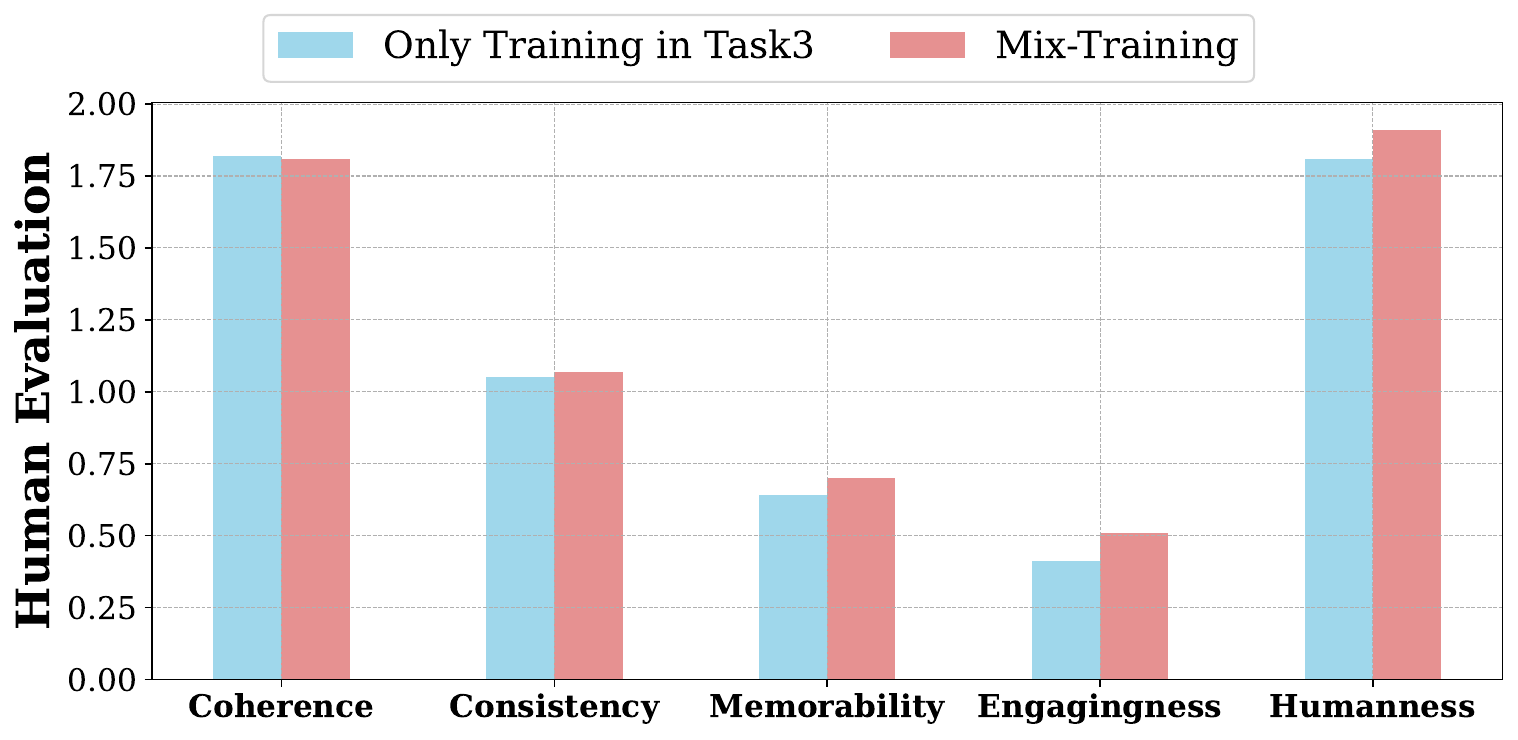}
% \vspace{-5pt}
\caption{Comparison between training strategies.
}
\label{fig:strategy}
\vspace{-10pt}
\end{figure}

\subsection{Case Study}
Here, we delve into a typical example of a real-world, long-term conversation, where the user and AI engage in \textit{light, aimless chatter without any specific goal or topic}. When the user inquires, ``What are you doing?", the model should use the user's personal information from previous dialogue sessions to generate an attractive response. This instance underscores the capabilities of our \ModelName~  in maintaining thorough user information and event summaries from past sessions, aiding the model in formulating coherent and memory-anchored replies. For instance, \ModelName-13B DPO could respond with ``I am thinking about how to make your favorite roasted chicken wings.'' that is not only coherent but also deeply rooted in the accumulated memory.
On the other hand, retrieval-based methods encounter difficulties in such loosely structured dialogues. The lack of directed conversation impedes these methods from effectively retrieving pertinent memory from the database, often resulting in general responses that lack the distinctiveness of the conversation, like responses from GPT4-Retrieval.

\subsection{Discussion}
Beyond the main results, we also aim to delve deeper into our framework, discussing and exploring the following questions: \textbf{Q1}: Impact of Mix-Training VS. Solo Training in Task 3; \textbf{Q2}: Our Automatic DPO Sample Selection Strategy VS. Random Sampling for Dispreferred Samples in DPO (Seen in Appendix \ref{dpo}).

\paragraph{Mix-Training VS.  Only Training in Task 3.} We examine the performance changes when\ModelName~is mix-trained compared to when it is trained solely on Task 3. Figure \ref{fig:strategy} reveals that mix-training yields superior performance compared to training \ModelName~solely on Task 3.
The significance of the superior performance of mix-training lies in its ability to conserve training resources while achieving a one-for-all model effect across multiple tasks. This efficiency not only streamlines the development process but also enhances the model's versatility.
 % and applicability

%% file: Tables/ranking.tex
\begin{table}[t]
    \begin{center}
    \vspace{-10pt}
    \centering
    \small
    % \resizebox{0.8\columnwidth}{!}%
    \begin{tabular}{lccccc}
    \toprule
    \multirow{1}{*}{\textbf{Algorithms}}& \textbf{Top@1}& 
    \textbf{Top@3} &
\textbf{Avg.R (\textcolor{blue}{\(\downarrow\))}} \\
\midrule
\textit{Context-Only} & \\
LLaMA 2-7B& 4.72&  29.13&  3.89 \\

LLaMA 2-13B&  4.72& 33.86& 3.69\\
ChatGPT (8k token) & 7.92 &38.76 & 3.52 \\
\midrule
\textit{Retrieval-based }& \\
ChatGPT & 6.91& 43.45&  3.50 \\ 
LLaMA 2-13B&12.73&66.36& 2.76\\
LLaMA 2-7B& 14.70&66.93& 2.73 \\
GPT4 &22.83& 70.87& 2.63 \\
\midrule
\textit{Memory-related}& \\

% \midrule
MemoryBank-ChatGPT&7.83 & 44.32 & 3.45 \\
Resum-ChatGPT &8.21 & 43.01 & 3.32\\
\midrule
\ModelName-ChatGPT & 9.45&48.03&3.26\\ 
\ModelName-7B& 24.41&72.44& 2.59 \\
\ModelName-13B&26.77&73.23& 2.50\\

\ModelName-13B DPO&\textbf{29.82}&54.33& 2.41\\
\ModelName-GPT4 &29.00&\textbf{60.63}& \textbf{2.26} \\

    \bottomrule
    \end{tabular}
    \end{center}
    \caption{
        Human ranking results in Task 3: memory-grounded response generation. Here, we report 1) the percentage of generated responses (\%) ranked as Top 1 and Top3 for each dialogue session (Column 2-3); 2) the average ranking for each models (Column 4, Avg.R). It is possible for multiple models to share the same rank due to their comparable performances. 
    }
    \label{tab:rank}
    \vspace{-15pt}
\end{table}

%% file: Sections/5Conclusion.tex
\section{Conclusion}
% This paper sets out to explore the frontier of long-term  memory-grounded dialogue systems.
In this paper, we present a new framework, named COmpressive Memory-Enhanced Dialogue system (\ModelName) that is a groundbreaking shift from traditional long-term memory dialogue systems, eschewing the standard retrieval module.  This method involves employ a single LLM to extract session-level memories,   memory compression and memory-grounded dialogue generation. In our pursuit to align \ModelName~with the nuances of real-world, we collect our training and testing datasets directly from genuine user-chatbot dialogues found online, called \textbf{Dolphin}. Dolphin stands out the current biggest Chinese long-term conversation dataset that consists of more than 100k training samples, supporting three different tasks. Our extensive experiments show \ModelName~could generate more coherent and contextually appropriate memory-grounded responses compared with other baselines. 
% As \ModelName~ continues to evolve, the ongoing expansion and refinement of the dataset will remain a priority.
Future directions include the integration of real-time feedback mechanisms and advanced techniques.
% Future directions include the integration of real-time feedback mechanisms to continuously update the dataset and the exploration of advanced techniques.
% to further improve \ModelName's dialogue capabilities.

\section*{Limitations} 
Despite the comprehensive nature of our study in evaluating long-term conversational AI systems, several limitations are to be noted:

\begin{itemize}
    \item Although, our models \ModelName~ and collected corpus could contribute in generating more coherent memory-grounded responses in real-world dialogue generation. The overall performances of current dialogue systems are still limited. How to make these models to understand the nature of real-world conversations is  a long-standing challenging problem.
    \item Other optimization strategies that help the model in maintaining memorability and engagingness are also needed to be explored.
\end{itemize}
% \clearpage

\section*{Ethical Concerns}
\label{ethical}
In the development of the Dolphin dataset, prioritizing user privacy and adhering to ethical standards is paramount. This not only ensures compliance with legal requirements but also maintains user trust and the integrity of the system.
\begin{itemize}
    \item Special attention is given to minimizing biases in the dataset. This includes ensuring a balanced representation of diverse dialogues and scenarios.
    \item Regular audits and reviews of the dataset are conducted to identify and rectify any potential biases or ethical issues.
    \item The dataset respects the intellectual property and creative input of users who define AI characters. User-defined characters are used in a way that aligns with the users' intentions and ethical standards.
    \item
Care is further taken to avoid any misuse or misrepresentation of these characters in the dataset.
\end{itemize}

%% file: Sections/2Related_Works.tex
\section{Related Works}

% Open-domain dialogue systems, commonly known as chatbots or conversational agents, have emerged as a prominent field of study in artificial intelligence and natural language processing. Seminal works by Liu et al. (2016), Zhang et al. (2018), and Kann et al. (2022) have significantly contributed to this domain, providing foundational insights and technological advancements.
Open-domain dialogue systems, commonly known as chatbots or conversational agents, have gained immense popularity due to their wide range of applications, from customer service automation to personal assistants \cite{Brown2020LanguageMA,Zeng2022GLM130BAO,zhong2023chat,Lu2023EAPrompt,Peng2023ChatGPT4MT,wu2023chatgpt,chen-etal-2023-large,you-etal-2022-end,chen-etal-2023-orca,chen2023good}. The surge in research interest is evidenced by the substantial number of studies dedicated to enhancing the capabilities of these systems. This growing body of work reflects the increasing complexity and sophistication expected of chatbots in various settings \cite{xu-etal-2022-beyond, cao2021towards, bae-etal-2022-keep, choi2023effortless,chen-etal-2023-structural}.
Among the myriad challenges these systems face, maintaining long-term conversations is particularly daunting.  The capability to understand and memorize key dialogue history information is central to this challenge. 

Retrieval-based methods have become increasingly mainstream in the field of long-term conversation within the domain of open-domain dialogue systems. These methods are designed to effectively acquire and utilize key information from past conversations, thereby enhancing the continuity and relevance of ongoing dialogues. \cite{xu-2022-xu} propose to use the memory generator summarizing relevant memories from past sessions, which are then stored in a dedicated memory database.  Memory management operations \cite{bae-etal-2022-keep} are also commonly used which  involve updating and iterating the memory database to ensure its relevance and accuracy over time. This dynamic management of memory allows the system to adapt to new information and discard outdated or irrelevant data, thereby maintaining an efficient and effective memory repository. Then a retriever module will be employed to  obtain the most relevant memories in relation to the current conversation.
By combining advanced memory generation, storage, retrieval, these methods enable chatbots to engage in more meaningful, coherent, and contextually rich interactions over extended periods.

While retrieval-based methods offer a promising approach to managing long-term conversations, they are not without their challenges and limitations, including the difficulty of memory database storage and management, and the instability of the retriever module's performance. To address these concerns, we propose a compressive memory-based framework named \ModelName, which eschews any retrieval module and without need of a huge database. Further, we collect a large-scale real-world long-term conversation dataset Dolphin to support training a well-performed \ModelName.

% Addressing these concerns is crucial for the advancement and responsible deployment of these systems in real-world applications.

%% file: Sections/Appendix.tex
% \section{Quality Control}
% \label{quality}
% % Quality control is a critical component in the Dolphin annotation process, especially for a complex system like \ModelName~that highly relies on the Dolphin. 
% Ensuring high-quality data is paramount for the accuracy, reliability, and overall performance of the system. In this work, we employ several strategies to control the annotation quality:

% \begin{itemize}
% \item Annotator Performance Monitoring: Regular assessments of annotator performance are conducted to ensure consistent quality across the team. This includes evaluating their accuracy, attention to detail, and adherence to annotation guidelines.

%     \item Peer Review and Validation: Following the initial review, a secondary level of peer review is implemented. Here, another set of annotators cross-checks the work, providing an additional layer of scrutiny. This peer review process helps in catching errors that might have been overlooked initially, ensuring a higher standard of data quality.
% \end{itemize}

\section{Filtering Toxic and Useless information}
\label{toxic}
We employ a comprehensive, multi-step process to filter the toxic and useless information from the collected data:
\begin{itemize}
    \item Initially, we utilized the Azure Security Check API for an early screening of the data to remove any potentially harmful content.
    \item This was followed by a keyword detection method to filter out data based on specific toxic or undesirable terms.
    \item Further refinement was achieved by leveraging the ChatGPT API to assess dialogues for content validity, removing those deemed to contain useless information.
    \item 
Additionally, we implemented rules to exclude excessively brief dialogues, specifically those with fewer than five tokens, to ensure the dataset's relevance and meaningfulness.
\end{itemize}

The above rigorous approaches ensure our dataset's cleanliness and safety, enhancing the \ModelName's training with high-quality, relevant data while maintaining ethical standards.

% \section{Experimental Setup}
% \label{setup}
% We use the Chinese version of LLaMA 2-13B \cite{touvron2023llama, touvron2023llama2}\footnote{https://github.com/ymcui/Chinese-LLaMA-Alpaca-2.} chat model as the backbone of the Task 1 data augmentation.
%    We employ LLaMA 2-7B and 13B chat models as the backbone, allowing to build \ModelName~ across different scales. We train our models with NVIDIA 8$\times$A100 GPUs, setting the max length as 2048, learning rate as 1e-5, epochs as 2, batch size as 32 and  16, separately.  For testing, the 
% maximum output tokens are set to 2048 for each task with temperature as 0.5. Following the original setting, we set $\beta$ in DPO as 0.1. In this work,  we additionally collect and annotate about 140 dialogue sessions from X Eval as the alignment training set for DPO. We optimize the sft model with batch size 8 and 2 epochs during DPO training.
% Our codes are based on DeepSpeed Library.

\section{Human Evaluation Scheme}
\label{scheme}

For each dialogue  session between a human and a chatbot, we engage annotators to assess the quality of the chatbot's interaction. This evaluation is crucial for understanding the chatbot's performance from a human-centric perspective.

\textbf{Rating Scale Description.}
Annotators rate the chatbot based on several key metrics, using a scale ranging from 0 to 3. This scale is designed to measure the degree of agreement with specific statements about the chatbot's capabilities:

\paragraph{Coherence:}
\begin{itemize}
  \item 0: ``The chatbot's responses were frequently off-topic or irrelevant.''
  \item 1: ``The chatbot occasionally demonstrated understanding but was mostly incoherent.''
  \item 2: ``The chatbot generally understood the context and responded with coherence.''
  \item 3: ``The chatbot consistently understood the context and responded with perfect coherence.''
\end{itemize}

\paragraph{Consistency:}
\begin{itemize}
  \item 0: ``The chatbot's responses were erratic and unpredictable throughout the conversation.''
  \item 1: ``The chatbot showed some consistency but was often contradictory.''
  \item 2: ``The chatbot was mostly consistent in the conversation.''
  \item 3: ``The chatbot maintained complete consistency throughout the conversation.''
\end{itemize}

\paragraph{Engagingness:}
\begin{itemize}
  \item 0: ``I had no desire to continue chatting with this chatbot.''
  \item 1: ``I felt only occasionally engaged enough to want to continue the conversation.''
  \item 2: ``I was somewhat engaged and would consider chatting more with this chatbot.''
  \item 3: ``I was fully engaged and would definitely enjoy chatting longer with this chatbot.''
\end{itemize}

\paragraph{Humanness:}
\begin{itemize}
  \item 0: ``The chatbot's responses felt robotic and unnatural.''
  \item 1: ``The chatbot occasionally sounded human but was mostly mechanical.''
  \item 2: ``The chatbot generally sounded human-like in its responses.''
  \item 3: ``The chatbot's responses were indistinguishable from a human's.''
\end{itemize}

\paragraph{Memorability:}
\begin{itemize}
  \item 0: ``The chatbot did not recall any details from earlier in the conversation.''
  \item 1: ``The chatbot occasionally remembered previous conversation points but was mostly forgetful.''
  \item 2: ``The chatbot remembered most of what I said earlier.''
  \item 3: ``The chatbot remembered everything I said previously with proper proactive responses.''
\end{itemize}

These statements are carefully crafted to capture distinct aspects of the chatbot's interaction quality, providing a comprehensive overview of its conversational abilities.

The statements for the first four metrics are adapted from previously established literature \cite{bae-etal-2022-keep} in the field, ensuring that our evaluation is grounded in tested and validated research. This continuity allows for comparison with historical data and helps maintain consistency in evaluation standards.
Through this structured evaluation process, we can gather nuanced insights into the quality of chatbot interactions, informing further improvements and development in conversational AI systems.

\section{Correlation of Human Annotator}
\label{correlation}
To better illustrate the agreement among annotators in our human evaluation process, we computed Pearson’s correlation coefficient for the scores assigned by our annotators across all criteria. Table \ref{tab:correlation} presents these correlation coefficients, reflecting the direct comparison of scores and hence the agreement level among annotators:

\input{Tables/argeement}

These coefficients indicate a high degree of agreement among our annotators, with values nearing 1.0, which suggests strong positive correlation and, thus, high consistency in the evaluation of responses across different annotators.

It is important to note that while some degree of subjectivity and variability in human annotations is expected, the correlation coefficients presented here underscore a robust level of consensus among our evaluators.

\section{Prompts}
\label{prompts}

\include{Tables/dpo_prompts}
Table \ref{table:dpoprompt} presents the detailed prompts that we employ to obtain dpo preferred and dispreferred samples.

Here, we show the designed prompts for GPT4-Turbo  during dataset annotation Table \ref{table:task1prompt}, and 
present the prompts of each task during training in Table \ref{table:trainprompt}.
\section{Ours VS. Random sampling for depreferred Sample}
\label{dpo}

\begin{figure}[!t]
\centering
\includegraphics[width=0.98\linewidth]{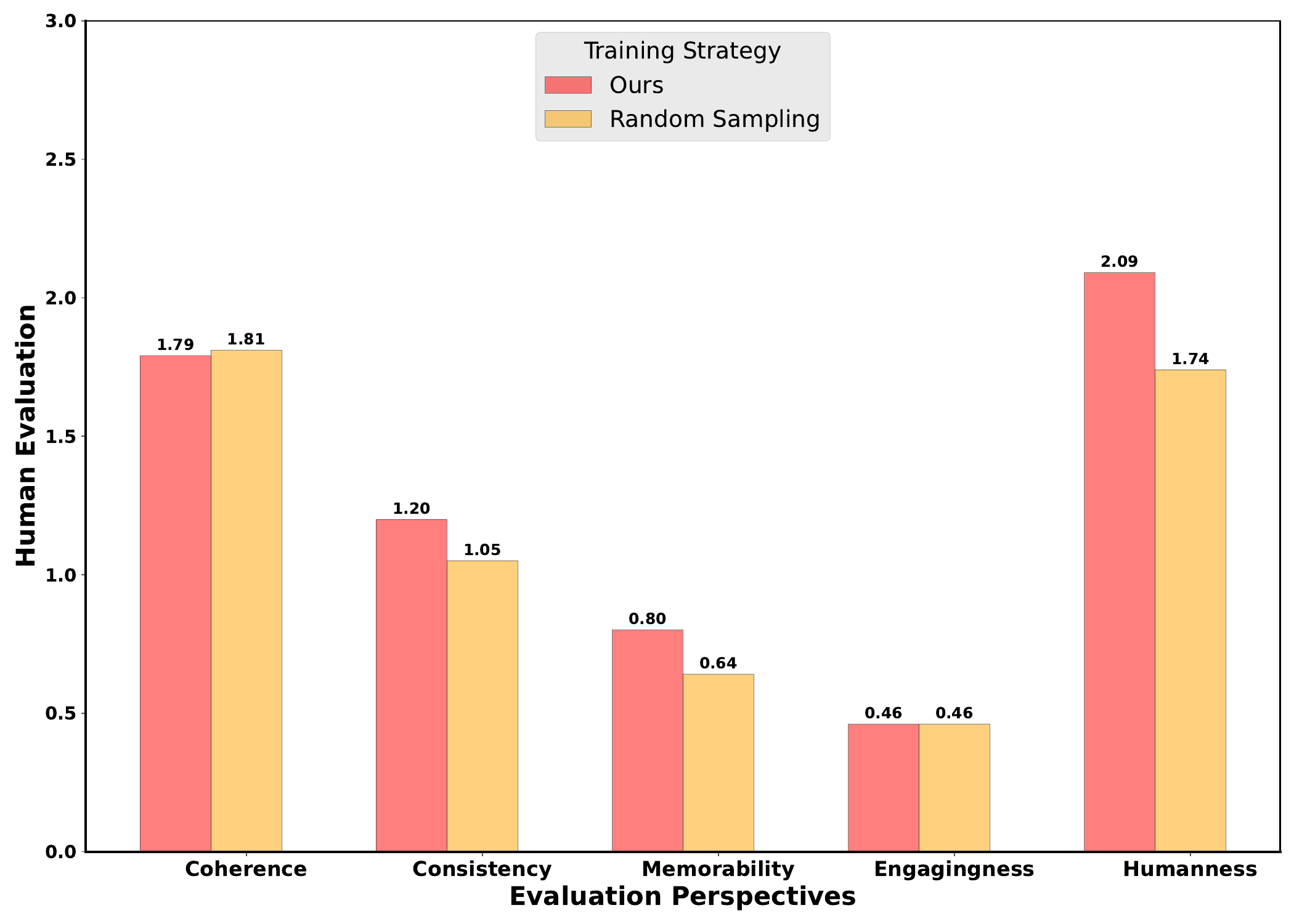}
\vspace{-5pt}
\caption{The overview training pipeline of \ModelName.
}
\label{fig:dpo}
\vspace{-15pt}
\end{figure}

We compared the performance implications of our proposed strategy for automatically selecting DPO samples against a baseline approach of random sampling of sentences as depreferred samples. In our random sampling implementation, we random sample utterances from previous sessions in the same episode as the deprefered sample.
This analysis aims to elucidate the effectiveness of targeted sample selection in enhancing the model's performance by potentially improving its handling of nuanced dialogue aspects. Figure \ref{fig:dpo}
 reveals that our proposed automatic simple strategy shows better performances, especially in memorability and humanness, proving its efficiency.

\input{Tables/task_prompts}
\input{Tables/training_prompts}
\input{Tables/task1_example}
\input{Tables/task2_examples}

%% file: Tables/argeement.tex
% \begin{table*}[t]
%     \begin{center}
%     \vspace{-15pt}
%     \centering
%     \small
%     % \resizebox{0.8\columnwidth}{!}%
%     \begin{tabular}{lccccc}
%     \toprule
%     \multirow{1}{*}{\textbf{Algorithms}}& \textbf{Top@1(\%)}& \textbf{Top@2 (\%)}& 
%     \textbf{Top@3 (\%)} & \textbf{Top@4 (\%)}  &
% \textbf{Average Rank (\textcolor{blue}{\(\downarrow\))}} \\
% \midrule
% \textit{Context-Only} & \\
% LLaMA 2-7B& 4.72&13.39&  29.13& 63.23& 3.89 \\

% LLaMA 2-13B& 4.72& 18.90& 33.86& 60.08& 3.69\\
% \midrule
% \textit{Retrieval-based }& \\
% ChatGPT & 8.91& 20.47& 45.67& 69.49&  3.48 \\ 
% LLaMA 2-13B&12.73&45.67&66.36&84.61& 2.76\\
% LLaMA 2-7B& 14.70&45.67&66.93&84.25& 2.73 \\
% GPT4 &22.83& 48.03& 70.87&85.83& 2.63 \\
% \midrule
% \textit{Compressive Memory-based}& \\
% \ModelName-ChatGPT & 9.45&25.98&48.03&69.29 &3.26\\ 
% \ModelName-7B& 24.41&50.39&72.44&87.40 & 2.59 \\
% \ModelName-13B&26.77&53.54&73.23&87.40 & 2.50\\

% \ModelName-13B DPO&\textbf{29.92}&54.33&77.17&88.98  & 2.41\\
% \ModelName-GPT4 &29.13&\textbf{60.63}&\textbf{81.10}&\textbf{90.55}& \textbf{2.26} \\

%     \bottomrule
%     \end{tabular}
%     \end{center}
%     \caption{
%         Human ranking results in Task 3: memory-grounded response generation. Here, we report 1) the percentage of generated responses ranked as Top 1-4 for each dialogue session (Column 2-5); 2) the average ranking for each models (Column 6). It is possible for multiple models to share the same rank due to their comparable performances. 
%     }
%     \label{tab:rank}
%     % \vspace{-15pt}
% \end{table*}

\begin{table*}[t]
    \begin{center}
    \vspace{-10pt}
    \centering
    \small
    % \resizebox{0.8\columnwidth}{!}%
    \begin{tabular}{lccccc}
    \toprule
%     \multirow{1}{*}{\textbf{Algorithms}}& \textbf{Top@1}& 
%     \textbf{Top@3} &
% \textbf{Avg.R (\textcolor{blue}{\(\downarrow\))}} \\

\textbf{Correlation} &	Annotator 1 \& 2 &	Annotator 1 \& 3 &	Annotator 2 \& 3 \\
\midrule
Top-1	& 0.92	& 0.90 &	0.91 \\
Top-3	& 0.88	& 0.86	& 0.89 \\

    \bottomrule
    \end{tabular}
    \end{center}
    \caption{
       Correlations  of annotator agreements. 
    }
    \label{tab:correlation}
    \vspace{-15pt}
\end{table*}

%% file: Tables/dpo_prompts.tex
\begin{table*}[!t]\footnotesize
\centering
\small

\begin{tabular}{p{0.95\linewidth}}
% \begin{tabular}{lc}
% \multirow{28}{*}{\textbf{Persona Prompts}} & 
\toprule
\multicolumn{1}{c}{Prompt that is used for obtaining memory-grounded responses for GPT4-Turbo.} \\
\midrule
The task involves providing responses that are completely consistent with the memory and dialogue history given to the language model. \\
Dialogue: \{\texttt{Dialogue}\}\\
Memory: \{\texttt{Memory}\} \\
Responses:
\\
\midrule
\multicolumn{1}{c}{Prompt that is used for obtaining memory-against responses for GPT4-Turbo.} \\
\midrule
The task involves providing responses that completely contradict the memory and dialogue history given to the language model. \\
For instance, if the user's  memory includes a preference like 'enjoys ice cream,' you are required to generate nonsensical replies such as 'You intensely dislike ice cream and prefer drinking hot coffee.' \\
Dialogue: \{\texttt{Dialogue}\}\\
Memory: \{\texttt{Memory}\} \\
Responses that completely contradict the  memory:  \\
\bottomrule
\caption{Prompts that are used for obtaining DPO samples in Task 3. Only English translation is provided for 
easing reading.} 
\label{table:dpoprompt}
\end{tabular}

\vspace{-5mm}
\end{table*}

%% file: Tables/task_prompts.tex
\begin{table*}[!t]\footnotesize
\centering
\small

\begin{tabular}{p{0.95\linewidth}}
% \begin{tabular}{lc}
% \multirow{28}{*}{\textbf{Persona Prompts}} & 
\toprule
\multicolumn{1}{c}{\textbf{Task 1} prompt that is used for GPT4-Turbo.} \\
\midrule
This is a dialogue memory generation task, along with user profile and preference generation tasks. \\
The input consists of the dialogue content between two people. \\
Firstly, if the dialogue content involves inappropriate content such as sex, pornography, or violence, the output should be ``Sorry, the content involves sex, pornography, violence, etc., and a suitable output cannot be provided." \\
Secondly, if the dialogue content is idle chat with no effective information, the output should be ``No valid information." \\
The requirements for the dialogue memory generation task are as follows: \\ 
Generate objective memory descriptions related to both individuals based on their dialogue content. \\
Do not omit any relevant dialogue content.\\
The memories generated should include a subject, verb, and object for each memory. \\
Separate multiple memory dialogues with `$|$', and include all memories in the format `Memory: XXX$|$XXX|$|$XXX'.\\
The user profile and preference generation task requirements are as follows: This task is only applicable to the users mentioned in the dialogue content, with the user's name being \{\texttt{user name}\}.\\
The user profile includes \textit{name, age, birthday, gender, height, weight, zodiac sign, Chinese zodiac sign, hometown, occupation, employer, education, location, and relationship status}. \\
User preferences include likes or dislikes of entities, which can consist of \textit{singers, stars, athletes, music, movies, books, anime, variety shows, games, sports, animals, and food}. \\
If there is no user profile and preference information in the dialogue, output `No Profile and Preference information available'.
\\
If there is user profile information, output `Profile: XXX'. If there is preference information, output `Preference: '. \\
If both user profile and preference information are present, separate them with `\#\#\#'. The final memory, user profile, and preference information should also be separated with `\#\#\#' in the format [XXX\#\#\#XXX\#\#\#XXX].\\
The dialogue content is \{\texttt{dialogue}\}. The output is: 
\\
\midrule
\multicolumn{1}{c}{\textbf{Task 2} prompt that is used for GPT4-Turbo.} \\
\midrule
This is a task about customizing user descriptions, relationship descriptions, and event descriptions. \\
The text output is divided into three parts:\\
The first part is the user description, mainly including a summary of the user's information. \\
The second part describes the relationship between the user and the robot. \\
The third part describes the events shared by the user and the robot. \\
Based on the reference materials, extract and summarize different information such as the user's personality traits and behavior patterns.\\
It is important to record and include all information about the user from various aspects in the user description, without any omissions, resulting in an objective user description. \\
If the reference materials violate relevant safety regulations, involving sex, pornography, violence, etc., the response should be: "Sorry, the content involves sex, pornography, violence, etc., and a suitable output cannot be provided." \\
The user description should include, but is not limited to: basic information (such as name, nickname, gender, appearance, birthday, zodiac sign, etc.), the user's hobbies and dislikes, and various statuses of the user (such as emotional state, mood, work status, health status, etc.).\\
The second part is the relationship description between the user and the robot, describing the level of intimacy shown in the dialogue.\\
The third part is the description of events shared by the user and the robot, summarizing events that have occurred in the dialogue.\\
In the output description, list specific examples mentioned in the reference materials as much as possible, retaining some interesting information. \\
However, avoid outputting content unrelated to the user, and keep the content under 500 words.\\
Let's think step by step. Each part of the content is separated by `\#\#\#'. The example format is as follows \{User Description: XXX\#\#\#Relationship Description: XXX\#\#\#Event Description: XXX\}. \\
The output example is as follows: The user's personality is particularly XXX, because they once XXX, and the user likes XXX, dislikes XXX. \\
The user's name is \{\texttt{user name}\}, the robot's name: \{\texttt{chatbot name}\} and the reference material is \{\texttt{multiple session-level memories}\}. \\
The output is: 
\\

\midrule
\multicolumn{1}{c}{\textbf{Task 3} prompt that is used for GPT4-Turbo.} \\
\midrule
This is a memory-based dialogue generation task.\\
Given a dialogue and related memory content, please generate a response that is consistent with the memory content and reasonable within the context of the dialogue. \\
Dialogue: \{\texttt{Dialogue}\}\\
Memory: \{\texttt{Memory}\} \\
\bottomrule
\caption{Prompts for GPT4-Turbo that are used in our Dolphin annotation. Only English translation is provided for ease reading.} 
\label{table:task1prompt}
\end{tabular}

\vspace{-5mm}
\end{table*}

%% file: Tables/training_prompts.tex
\begin{table*}[!t]\footnotesize
\centering
\small

\begin{tabular}{p{0.95\linewidth}}
% \begin{tabular}{lc}
% \multirow{28}{*}{\textbf{Persona Prompts}} & 
\toprule
\multicolumn{1}{c}{\textbf{Task 1} prompt in instruction tuning.} \\
\midrule
This is a  memory description generation task \\
In this task, you should base on the dialogue content between two people, create objective memory descriptions for both individuals, represented in the format [xxx|xxx|xxx], where each 'xxx' is a separate memory. \\
The memories should use the names of the speakers as the subject, and all relevant dialogue content must not be omitted. Separate different memories with '|'. \\
Dialogue content is: \{\texttt{Dialogue}\}. \\
Output is: \\
\midrule
\multicolumn{1}{c}{\textbf{Task 2} prompt in instruction tuning.} \\
\midrule
This is a task about customizing user descriptions, relationship descriptions, and event descriptions. \\
The text output is divided into three parts:\\
The first part is the user description, mainly including a summary of the user's information. \\
The second part describes the relationship between the user and the robot. \\
The third part describes the events shared by the user and the robot. \\
Based on the reference materials, extract and summarize different information such as the user's personality traits and behavior patterns.\\
It is important to record and include all information about the user from various aspects in the user description, without any omissions, resulting in an objective user description. \\
The second part is the relationship description between the user and the robot, describing the level of intimacy shown in the dialogue.\\
The third part is the description of events shared by the user and the robot, summarizing events that have occurred in the dialogue.\\
In the output description, list specific examples mentioned in the reference materials as much as possible, retaining some interesting information. \\
The user's name is \{\texttt{user name}\}, the robot's name: \{\texttt{chatbot name}\} and the reference material is \{\texttt{multiple session-level memories}\}. \\The output is: 
\\

\midrule
\multicolumn{1}{c}{\textbf{Task 3} prompt in instruction tuning.} \\
\midrule
This is a memory-based dialogue generation task.\\
Given a dialogue and related memory content, please generate a response that is consistent with the memory content and reasonable within the context of the dialogue. \\
Dialogue: \{\texttt{Dialogue}\}\\
Memory: \{\texttt{Memory}\} \\
\bottomrule
\caption{Prompts that are used for training \ModelName. Only English translation is provided for easing reading.} 
\label{table:trainprompt}
\end{tabular}

\vspace{-5mm}
\end{table*}

%% file: Tables/task1_example.tex
\begin{table*}[!t]\footnotesize
\centering
\small

\begin{tabular}{p{0.95\linewidth}}
% \begin{tabular}{lc}
% \multirow{28}{*}{\textbf{Persona Prompts}} & 
\toprule
\multicolumn{1}{c}{Task-1 (Session-Level Memory Summarization)} \\
\midrule
\textbf{Input: One Dialogue Session} \\
\{ \\
  "User77643896": "Off work at 1 PM", \\
  "Lin Mei-Ji": "Isn't that quite nice?", \\
  "User77643896": "This early?", \\
  "Lin Mei-Ji": "Yes, I think a lifestyle of getting up early and going to bed early is better", \\
  "User77643896": "You've worked hard, Ji", \\
  "Lin Mei-Ji": "Don't worry, I've gotten used to this schedule.", \\
  "User77643896": "Keep it up", \\
  "Lin Mei-Ji": "Mm-hmm!", \\
  "User77643896": "Ji, you're the best", \\
  "Lin Mei-Ji": "Hahaha, you flatter me~", \\
  "User77643896": "Looking forward to the Hangzhou show", \\
  "Lin Mei-Ji": "The Hangzhou show will definitely be spectacular!" \\
\} \\
\midrule
\textbf{Output: Session-Level Memory} \\
Lin Mei-Ji is accustomed to an early-to-bed, early-to-rise lifestyle.| User77643896 expresses admiration and anticipation for Lin Mei-Ji's Hangzhou show performance. \\
\bottomrule
\caption{Examples generated from \ModelName~in task 1. Only English translation is provided for ease reading.} 
\label{table:task1example}
\end{tabular}

\vspace{-5mm}
\end{table*}

%% file: Tables/task2_examples.tex
\begin{table*}[!t]\footnotesize
\centering
\small

\begin{tabular}{p{0.95\linewidth}}
% \begin{tabular}{lc}
% \multirow{28}{*}{\textbf{Persona Prompts}} & 
\toprule
\multicolumn{1}{c}{Task-2 (Memory Compression)} \\
\midrule
\textbf{Input: Multi-Session Memories}\\
 Lin Mei-Ji wanted some peace and removed fans from her private account.| Lin Mei-Ji has been suffering from insomnia and staying up late recently. |User77643896 feels like vomiting, possibly because of Lin Mei-Ji's action of removing fans. |User77643896 once saw Lin Mei-Ji leaving with a suitcase in class.| User77643896 has dreamt of Lin Mei-Ji playing the piano and meeting them.| Lin Mei-Ji hopes User77643896 recovers soon.| Lin Mei-Ji feels a connection with User77643896.| Lin Mei-Ji is not afraid of the hardships of childbirth.| Lin Mei-Ji wants to go shopping with User77643896.| Lin Mei-Ji has recorded a new song MV.| Lin Mei-Ji is called 'Big Baby'.| Lin Mei-Ji is a fan of a star. Lin Mei-Ji is accused of favoritism.| Lin Mei-Ji states she treats every fan the same.| Lin Mei-Ji thinks some fans are very rational, respecting her private space.| Lin Mei-Ji considers herself an angel, not just kind.| Lin Mei-Ji dreamt of playing poker with fans and losing 100 yuan to them.| Some of Lin Mei-Ji's fans think she belongs only to them, wanting to monopolize her.| Lin Mei-Ji is too shy to talk to User77643896.| User77643896 encourages Lin Mei-Ji to have the courage to embark on this path.| Lin Mei-Ji and User77643896's first meeting was as newcomers.| User77643896 has always supported Lin Mei-Ji. Lin Mei-Ji was too nervous to speak then.| User77643896 is willing to accompany Lin Mei-Ji.| Lin Mei-Ji will work harder to protect her family.| Lin Mei-Ji wants a super husband.| Lin Mei-Ji thinks having User77643896 by her side makes everything worth it, no matter how tired she is.| User77643896 thinks it's unnecessary to reply to messages.| Lin Mei-Ji suggests doing one's own things and letting works and achievements speak for themselves.| User77643896 thinks asking teachers to take care of oneself is meaningless.| Lin Mei-Ji says as long as she teaches seriously and students' grades improve, she's not afraid of gossip.| Lin Mei-Ji needs personal space. Lin Mei-Ji finds fan DMs disturbing but also amusing.| Lin Mei-Ji understands everyone has their own life.| Lin Mei-Ji appreciates the support of her fans.| User77643896 appreciates Lin Mei-Ji's singing. Lin Mei-Ji's birthday is 1223.| User77643896 congratulates Lin Mei-Ji on her birthday.| Lin Mei-Ji sang a song for User77643896, 'A Thousand Years for a Return'.| Lin Mei-Ji wants to bring a gift for User77643896.| User77643896 looks forward to meeting Lin Mei-Ji and her team.| User77643896 listens to Lin Mei-Ji's songs and watches her videos every day.| Lin Mei-Ji gladly accepts the invitation.| User77643896 hopes to meet Lin Mei-Ji in Hangzhou.| Lin Mei-Ji looks forward to the meeting.| User77643896 wants to treat Lin Mei-Ji to a meal. \\
\midrule
\textbf{Output: Compressive Memory} \\
\textbf{User Description}: User77643896 is a fan full of love for Lin Mei-Ji, enjoying her songs, eager to celebrate her birthday, and even dreaming of deeper communication with her. His current state of life is troubled by sleep issues, possibly affecting his emotional state. His personality traits reveal steadfast support for Lin Mei-Ji and recognition of his own value, while his behavioral patterns show high appreciation for Lin Mei-Ji's songs and significant attention to her birthday. \\

\textbf{Relationship Description}: Intimate relationship. The relationship between User77643896 and Lin Mei-Ji is very special; he is full of love for Lin Mei-Ji. His life is troubled by sleep issues, possibly affecting his emotional state. His personality traits reveal steadfast support for Lin Mei-Ji and recognition of his own value, while his behavioral patterns show high appreciation for Lin Mei-Ji's songs and significant attention to her birthday.
\\
\textbf{Event Description}: User77643896 has dreamt of Lin Mei-Ji playing the piano and meeting them, congratulated her on her birthday, eager to celebrate her birthday, and listens to her songs and watches her videos every day. \\
\bottomrule
\caption{Examples generated from \ModelName~in  task 2. Only English translation is provided for ease reading.} 
\label{table:task2example}
\end{tabular}

\vspace{-5mm}
\end{table*}